\documentclass[pdflatex,sn-mathphys-num]{sn-jnl}

\usepackage{graphicx}%
\usepackage{multirow}%
\usepackage{amsmath,amssymb,amsfonts}%
\usepackage{amsthm}%
\usepackage{mathrsfs}%
\usepackage[title]{appendix}%
\usepackage{xcolor}%
\usepackage{textcomp}%
\usepackage{manyfoot}%
\usepackage{booktabs}%
\usepackage{algorithm}%
\usepackage{algorithmicx}%
\usepackage{algpseudocode}%
\usepackage{listings}%
\usepackage[T1]{fontenc}
\usepackage{subcaption}
\usepackage{float}
\usepackage{tabularx}
\usepackage{graphicx,verbatim} 
\usepackage{multirow}
\usepackage{cleveref}
\usepackage[normalem]{ulem}

\usepackage{textcomp}
\usepackage{verbatim}
\usepackage[commandnameprefix=ifneeded,markup=underlined]{changes}

\theoremstyle{thmstyleone}%

\theoremstyle{thmstyletwo}%

\theoremstyle{thmstylethree}%

\begin{document}

\title[Article Title]{PICO: Projection-Informed Consistency Optimisation for 6DoF Surgical Tool Pose Estimation}

\author*[1]{\fnm{Lucy} \sur{Fothergill}}\email{sclef@leeds.ac.uk}

\author[2]{\fnm{Pietro} \sur{Valdastri}}

\author[2]{\fnm{Dominic} \sur{Jones}}
\author[1]{\fnm{Duygu} \sur{Sarikaya}}

\affil[1]{School of Computer Science, University of Leeds}

\affil[2]{School of Electronic and Electrical Engineering, University of Leeds}

\abstract{The abstract serves both as a general introduction to the topic and as a brief, non-technical summary of the main results and their implications. Authors are advised to check the author instructions for the journal they are submitting to for word limits and if structural elements like subheadings, citations, or equations are permitted.}

 \abstract{\textbf{Purpose:} Accurate 6 DoF pose estimation of surgical tools is critical for automation, robotic proprioception, and safe interaction with the tissue operated on. Kinematics-based approaches suffer from accumulated errors due to the cable-driven nature of robotic arms, while vision-based methods often rely on external markers or trackers. Although more recent vision-based advances have been proposed, these two-stage pose estimation methods often lack real-time robustness due to accumulated errors and computational overhead.

 \textbf{Methods:} We propose a novel end-to-end trainable model, PICO. Our model employs a multi-task learning architecture to predict segmentation and depth maps, alongside regression of translation and rotation parameters. We define two proxy tasks that enforce geometric consistency in both 2D and 3D spaces, improving accuracy and robustness. For this, we propose a projection loss, and a point-to-point loss.
 
 \textbf{Results:} We evaluate our method on the SurgRIPE dataset, benchmarking its performance against state-of-the-art approaches using standard 6DoF pose estimation metrics. Our results demonstrate consistently strong performance across all four subsets, specifically in rotation, ranking second even under occlusion. It also demonstrates comparable translational performance, remaining competitive, especially in occluded cases.

 \textbf{Conclusion:} PICO demonstrates the effectiveness of multi-task learning and geometry-aware proxy tasks for robust and reliable surgical tool pose estimation, especially in occluded scenarios, highlighting potential for future applications.}


\maketitle

\section{Introduction}\label{sec1}
Autonomous robotic surgery requires safe interaction with the surgical environment. A fundamental aspect of this is accurate positional knowledge of surgical tools, which is necessary for preoperative planning and the prevention of adverse tool-tool or tool-tissue interactions, and providing real-time feedback in response to changes in the environment. Current da Vinci\textregistered Surgical Systems (dVSS) estimate 6D pose of surgical tools with forward kinematic equations that rely on joint angle measurements of the robotic arm and known linkage parameters. Kinematics-based pose is not sensitive enough for autonomy, with discrepancies of up to 1.02mm, most noticeably in the end effector pose, arising due to cable-slack, pulley-cable friction and accumulation of errors across multiple joints \cite{dvrkcaveats}. Such inaccuracies are unacceptable in a surgical setting, as even small errors can lead to unintended incisions and damage to critical structures such as nerves or blood vessels. Therefore these limitations could be addressed with vision-based methods, which provide direct perception of the scene and have the potential to adapt to dynamic situations where conditions change, such as tissue deformation or unexpected events. The use of external markers and trackers in a surgical setting are impractical and potentially dangerous. Attaching markers to tools requires additional time-consuming procedures such as preoperative sterilisation which interrupts surgical workflow \cite{markerless1}. Additionally, methods which rely on external markers demand a clear line of sight between the camera and marker, which can not be guaranteed in a surgical environment due to occlusion from blood, tissue, smoke or other instruments \cite{markerless2}. For these reasons, markerless methods are preferable.
 
\begin{figure}[h!]
\centering
\includegraphics[width = \textwidth]{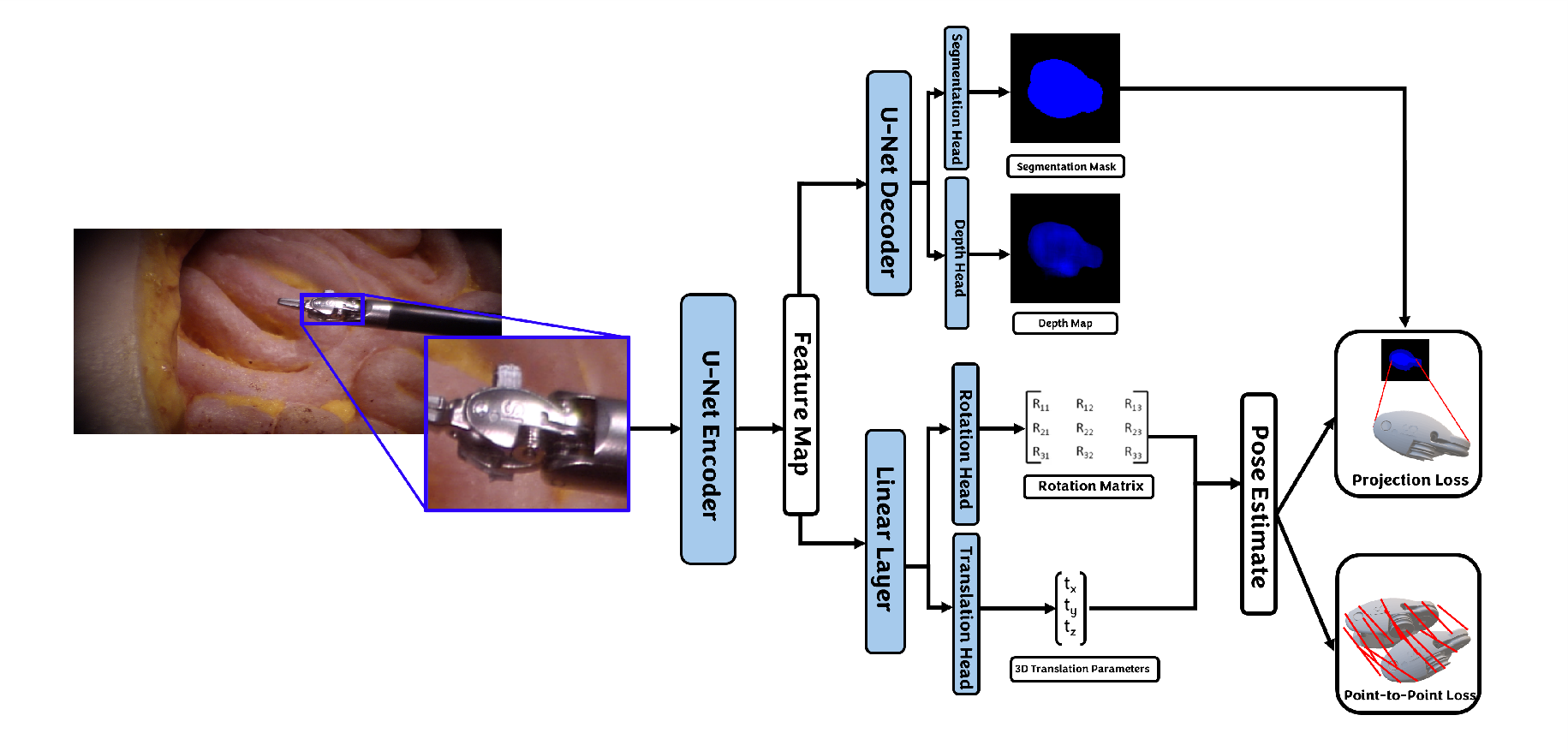}
\caption{\textbf{Proposed training framework} Given an input RGB image, the tool is first cropped, resized and fed into a U-Net style architecture with a ResNet-50 encoder backbone. The learned features are fed into a decoder and subsequently pass through separate depth and segmentation heads. Simultaneously, the encoder features are passed through linear layers and regression heads for pose estimation. The estimated pose is used to transform the 3D tool model, and both the 2D projection and 3D representation of the tool are minimised through two defined proxy tasks. All losses are optimized jointly with the multi task loss function.} \label{architecture}
\end{figure}

A major challenge in 6D surgical tool pose estimation is the lack of ground truth pose annotations for surgical images. To address this, many existing methods \cite{surgripechallenge, pvnet, occlusion-robust, weaklysupervised} avoid directly predicting the 6D pose and instead rely on multi-stage approaches that use more readily available geometric information to supervise intermediate predictions. These intermediate representations, such as keypoints are first predicted, and the 6D pose is then estimated using methods such as Perspective-n-Point (PnP), template matching, or rendering-based approaches. Alternatively, an initial pose can be estimated and subsequently refined iteratively. These multi-stage pipelines can struggle with real-time performance due to sensitivity to noise, accumulated errors and computational overhead. The accuracy depends on the quality of the 2D-3D correspondences which may be noisy or inaccurate. Moreover, if the initial pose estimation is inaccurate, the refinement step may not be able to fully correct it. Iterative refinement methods are also known to be slow. During inference, two-stage methods that first estimate 2D features and then apply Perspective-n-Point (PnP) or other refinement techniques may introduce delays. The iterative nature of pose refinement can be time-consuming at inference time, degrading the model’s performance for dynamic surgical settings. For methods relying on generating 2D-3D correspondences at inference time or rendered images, the complexity of the calculations can significantly impact inference speed.
In this work, we propose a novel end-to-end trainable model, PICO: Projection-Informed Consistency Optimisation for 6DoF Surgical Tool Pose Estimation. PICO leverages a multi-task learning architecture to simultaneously predict segmentation and depth maps, while also regressing translation and rotation parameters. We then define two proxy tasks that enforce geometric consistency in both 2D and 3D spaces, improving accuracy and robustness. 
First, we propose using a projection-based loss, which enforces geometric consistency between predicted pose and the predicted tool segmentation. This ensures that the model maintains spatial consistency through alignment of 2D observations and 3D model representations. The projection-based loss encourages the model to learn geometrically consistent representations, leading to a more accurate pose estimation. Discrepancies between the model's projection and predicted segmentation lead to penalisation in the loss function, which helps improve pose estimation accuracy during training, as the model learns to generate more accurate pose estimations for surgical tools. Similarly, we define a point-to-point loss, where the predicted and ground truth pose transformations are applied to the 3D model points, and the root mean squared error (RMSE) between corresponding points is minimised to promote 3D consistency. An overview of our method is shown in Figure \ref{architecture}. 

We evaluate our method on the SurgRIPE dataset \cite{surgripechallenge}, comparing its performance against existing approaches using standard 6DoF pose estimation metrics. Across all four test datasets, our end-to-end trainable model PICO ranks second in rotational accuracy even under occlusion. It also demonstrates comparable translational performance, remaining competitive, especially in occluded cases. These results highlight the effectiveness of our geometric consistency approach, enforced through our projection and point-to-point proxy tasks.

\section{Related Work}
\textbf{\textit{Object Pose Estimation}:}
State of the art 6D object pose estimation methods use the Co-op architecture with differing object detectors and number of hypotheses \cite{coop} \cite{bopleaderboard}. Pose estimation is carried out by a three-stage pipeline: rough estimation via semi-dense template matching, iterative refinement using dense optical flow and an optional pose selection module. Other high-performing methods \cite{picopose, megapose, gigapose, genflow} use an initial pose estimation step and a subsequent refinement technique,  such as predicting dense optical flow between real and rendered images, or a render-and-compare approach.Earlier networks that form the basis of applied 6D pose estimation approaches are GDR-Net \cite{GDRNet} and PVNet \cite{pvnet}. GDR-Net uses learned dense 2D-3D correspondences and surface-region attentions with a differentiable PnP refinement structure to regress 6D object pose, while PVNet predicts pixelwise unit vectors and a RANSAC voting scheme to obtain 2D keypoints. 6D pose is then acquired with a PnP solver. Effective generalisation of these methods to a surgical setting is challenging due to domain-specific characteristics like articulation and specular reflections from metallic tool surfaces \cite{spektor2025monocular}.

\noindent \textbf{\textit{Surgical Tool Pose Estimation}:}
Many existing methods for surgical tool pose estimation indirectly estimate pose, using multiple stages where intermediate geometric features are extracted first before 6D pose is obtained by some further algorithm. Hasan et al. propose a method that recovers 6D pose using algebraic geometry methods from segmentation \cite{intra2}, while Xu et al. estimate 2D tool keypoints and uses a PnP RANSAC-based solver to give final 6D pose \cite{xu2024occlusion}, and Rai et al. use RAFT-Stereo \cite{RAFTStereo} to estimate depth and combined with a segmentation mask for 6D pose estimation \cite{surgpose} \cite{foundationpose} \cite{sam6d}.
Few methods directly estimate the 6D pose of the surgical instrument. In \cite{spektor2025monocular}, Spektor et al. takes inspiration from GDR-Net, along with detected object crops and synthetic data to estimate 6D pose, object class and articulation angle of surgical tools during open surgery. Fan et al. introduce a framework that formulates pose estimation as a Markov Decision Process and solves using a reinforcement learning agent and a virtual articulated skeleton \cite{fan2024rlpose}. 
Xu et al. outline the methods that have been evaluated using the SurgRIPE dataset \cite{surgripechallenge}. ImFusion applies 2D object detection and SurfEmb \cite{surfemb} to generate surface embedding-based 2D-3D correspondences, then refines multiple pose hypotheses using a render-and-compare strategy. Other approaches adapt existing pose estimation methods, including PVNet with added data augmentation and a deglare algorithm, EfficientPose \cite{efficientpose} using DeepLabV3+ \cite{deeplabv3} for segmentation and ConvNeXt \cite{liu2022convnet} for multi-task pose prediction, and a ResNet-18 encoder with SwiftFormer for enhanced global context information extraction \cite{swiftformer}. However, multi-stage methods introduce additional computational overhead and are dependent on accuracy of intermediate predictions, so errors in segmentation, keypoint localisation or depth estimation can impact the final pose estimate. Additional inference steps used by methods that use correspondence generation, iterative render-and-compare refinement or PnP limit their suitablity for real-time surgical applications. In contrast, our proposed method directly predicts the 6D pose in a single step while maintaining geometric consistency.

\section{Methodology}
Given a cropped monocular RGB image as input, we develop a multi-task learning framework that estimates the 6 Degree-of-Freedom (DoF) pose of a surgical tool, where 6 DoF pose defines an object's position and orientation in 3D space. The network jointly predicts segmentation masks and depth maps, in addition to the pose, given by 3D rotation and translation parameters. The dataset provides ground truth rotations as 3x3 matrices. As such, we directly regress a 3x3 matrix which is projected onto the nearest valid rotation matrix in SO(3) using Singular Value Decomposition (SVD). This is used instead of alternate approaches, such as Euler angles, quaternions or 6D representations to ensure a singularity-free and unique parameterisation in 3D space and avoids unnecessary conversions, while translation is expressed relative to the image frame, with an (x,y) pixel coordinate and a depth z (distance to the camera), as proposed by Xiang et al. in \cite{posecnn} and used in \cite{efficientpose}. \\
First, features are extracted from the input image using a shared encoder. Following this, the feature map passes through a decoder network, then individual task-specific network heads for segmentation and depth prediction. For translation and rotation parameter estimation, the encoder features are first passed through a shared fully connected linear layer, before being fed into specific regression heads. The translation prediction task was split further into 2 subtasks: predicting the 2D (x,y) pixel coordinates of the tool joint in the image, and predicting the depth (z), relative to the camera. 
The model architecture is based on that of the U-Net \cite{unet}, with a ResNet-50 \cite{resnet50} based encoder for feature extraction from the input image. ResNet was used as the encoder backbone for the U-Net architecture, due to its computationally-efficient feature extraction capability, widespread use in medical image analysis and the availability of ImageNet pre-trained weights, which allow for effective transfer learning in this domain where data is limited. To generate a binary segmentation mask, the final output from the decoder is passed through a segmentation head, consisting of a 2D convolutional layer, producing a single-channel raw logits output with the same spatial dimensions as the input image. The same decoder output is fed into a depth head, which also applies a 2D convolution layer and an additional rectified linear unit (ReLU) activation layer. 
The features from the shared encoder network were also downsampled, flattened and then fed into a shared fully connected linear layer before passing into rotation and translation regression heads. Each head consists of an additional linear layer and an activation function. Tanh is the activation function used to constrain predicted elements of the rotation matrix to [-1,1], after which the matrix is orthogonalised to obtain a valid representation. The sigmoid function is used to constrain (x,y) translation values between [0,1] as these image pixel coordinates are normalised, while ELU is used for z translation prediction as it allows a wider range of depth values. The aim of performing the auxiliary tasks of segmentation and depth prediction is to improve pose estimation by encouraging the network to learn more spatially-aware representations. In particular, pseudo-depth maps provide additional 3D spatial supervision to the model, beyond the single predicted point, which helps to learn scale and perspective and leads to more robust feature representations.

\subsection{Loss Functions}
A multi-task loss function is used to train the network end-to-end. This function contains segmentation, depth, projection and point-to-point matching loss terms in addition to rotation and translation parameter losses.\\
\textbf{Segmentation Loss:} The Dice loss between the predicted and ground truth segmentation masks was computed and optimised to refine the model's segmentation performance.
\begin{equation*}
\mathcal{L}_{\text{seg}} =
1 -
\frac{
2 \left| \mathcal{M}_{\text{proj}} \cap \mathcal{M}_{\text{pred}} \right|
}{
\left| \mathcal{M}_{\text{proj}} \right|
+
\left| \mathcal{M}_{\text{pred}} \right|
}
\end{equation*}
\\
\textbf{Depth Loss:} For depth loss, the Scale-Invariant Loss, as proposed by Eigen et al. \cite{siloss} was used, which was specifically designed for monocular depth estimation. It includes two loss terms – the first a pixelwise error term that measures local differences and the second that ensures predictions retain correct relative depth relationships, even with variation in absolute scale. Pseudo-depth maps are created by transforming randomly sampled 3D tool meshes using the true tool pose and the projection equation $t = K^{-1} t_z \begin{bmatrix} o_x & o_y & 1 \end{bmatrix}^\top$ to give 2D pixel coordinates and a corresponding depth value which is assigned to the nearest pixel. Where there are multiple depth values for a given pixel, for example a point on the front and back face of the tool are assigned to the same image pixel, the minimum value is taken representing the point closest to the camera. To overcome sparsity and any holes in the map, zero-valued pixels are replaced with an average of neighbouring pixels. The Scale-Invariant loss term is calculated using the equation 
\begin{equation*}
\mathcal{L}_{\text{depth}} = \mathcal{L}_{\text{SI}}(y, \hat{y}) = \frac{1}{n} \sum_{i} (\log \hat{y}_i - \log y_i)^2
- \frac{1}{n^2} \left( \sum_{i} (\log \hat{y}_i - \log y_i) \right)^2
\end{equation*}
\\
\textbf{Projection Loss} The predicted pose transformation was applied to the a randomly sampled 3D point mesh of the tool and the transformed points were projected into the 2D plane using the camera intrinsics according to the equation $t = K^{-1} t_z \begin{bmatrix} o_x & o_y & 1 \end{bmatrix}^\top$. The resulting 2D mesh points were downsampled and the contour points of the concave hull generated a projected binary mask. This projected binary segmentation mask is used as a pseudo-ground-truth supervisory signal. The Dice loss between this and the predicted segmentation mask was calculated and used to ensure consistency and alignment between the tool's 2D image space appearance and 3D model space.
\begin{equation*}
\mathcal{L}_{\text{proj}} =
1 -
\frac{
2 \left| \mathcal{M}_{\text{proj}} \cap \mathcal{M}_{\text{pred}} \right|
}{
\left| \mathcal{M}_{\text{proj}} \right|
+
\left| \mathcal{M}_{\text{pred}} \right|
}
\end{equation*}
\\
\textbf{Point-to-Point Loss} Again, both the predicted and ground truth pose transformations were applied to the randomly sampled 3D model points, and the root mean squared error (RMSE) between corresponding points was minimised to promote 3D consistency.
\begin{equation*}
\mathcal{L}_{\text{p}} =
\left(
\frac{1}{N}
\sum_{i=1}^{N}
\left\lVert
R_{\text{pred}} \mathbf{X}_i + t_{\text{pred}}
-
\left( R_{\text{GT}} \mathbf{X}_i + t_{\text{GT}} \right)
\right\rVert_2^{2}
\right)^{\frac{1}{2}}
\end{equation*}
\\
\textbf{Rotation Loss} For improved rotation prediction, a geodesic loss function was used. Rather than standard Euclidean distance, which would treat rotation matrix elements as independent values, geodesic distance is a measure of the angular distance between two rotation matrices, which provides a more intuitive measure \cite{geodesic}. This loss is calculated by the equation \cite{geodesic2} 
\begin{equation*}
\mathcal{L}_{\text{R}} = d(R_s, R_{\text{GT}}) = \cos^{-1} \left( \frac{\operatorname{tr}(R_s^\top R_{\text{GT}}) - 1}{2} \right)
\end{equation*}
\textbf{Translation Loss} The root mean squared error (RMSE) between predicted and ground truth translations was minimised separately for the (x,y) and z components, in order to reduce Euclidean distance between the two points.
\begin{equation*}
\begin{aligned}
\mathcal{L}_{T_{xy}} &= \frac{1}{N} \sum_{i = 1}^{N} (  xy_{i} -  \hat{xy_{i}} ) ^{2} \\[6pt]
\mathcal{L}_{t_{z}} &= \frac{1}{N} \sum_{i = 1}^{N} (  z_{i} -  \hat{z_{i}} ) ^{2}
\end{aligned}
\end{equation*}
The total loss function was a sum of all these individual components: $$\mathcal{L}_{Total} =  \mathcal{L}_{seg} +  \mathcal{L}_{depth} +  \mathcal{L}_{proj} + \mathcal{L}_{p} +  \mathcal{L}_{R} +  \mathcal{L}_{T_{xy}}+  \mathcal{L}_{T_{z}}$$
Overall, the combination of these loss terms provide supervision across 2D and 3D domains. The segmentation and depth losses provide image-based supervision, while projection, point-to-point, rotation and translation losses directly encourage geometric and pose consistency. Together these losses encourage the network to learn accurate 6D poses while maintaining consistency between predicted tool appearances and underlying 3D geometry.
\pagebreak
\section{Experiments}
We evaluated the network on the SurgRIPE dataset \cite{surgripechallenge}, which included monocular RGB images of two surgical tools – the large needle driver (LND) and Maryland bipolar forceps (MBF). The dataset contains $1147$ LND and $1109$ MBF images, split $90$:$10$ for training and validation. For testing, there were two sets of images used for each tool – occluded and unoccluded.\\
We compared the performance of our network to the results of the SurgRIPE challenge \cite{surgripechallenge}, using the standard pose estimation metrics of translation error and rotation error, as well as an ADD metric. ADD measures the average distance between point clouds transformed by ground truth and predicted pose, and this metric evaluates the percentage of instances where the ADD is smaller than $10\%$ of the instrument’s diameter.
Before training, the ($x$,$y$) translation values were converted from camera frame to image frame, using the camera intrinsics, to give pixel coordinates, while the depth (distance to camera) remained the same.\\
 During training, the minimum and maximum $x$ and $y$ values of the segmentation mask were used to define the tool's bounding box in the image and the U-Net decoder was used to support auxiliary training tasks. At inference time, the decoder was discarded, reducing computational overhead and improving efficiency. Additionally, during inference, the bounding box the bounding box was predicted using a fine-tuned YOLOv5 object detection model, with IoU scores of 0.87, 0.86, 0.89 and 0.77 for LND, LND Occluded, MBF and MBF Occluded test datasets, respectively \cite{yolo}. In instances where multiple tools were present in the image, the detection with the highest level of confidence was taken as the object of interest. The tool's bounding box, with an additional 30 pixels added to each side, was then used to crop the image, which was resized to shape $(224,224)$ before being fed into the network.
The network was trained in PyTorch using a Google Colab environment with an NVIDIA Tesla T4 GPU with 16 GB of VRAM. A batch size of $8$ was used, for $50$ epochs with an Adam optimiser and a learning rate of $10^{-4}$. A stopping criteria was implemented so that training was terminated if the validation loss did not improve within $10$ epochs. 

\section{Results}\label{sec2}

\begin{table}[!h] 
\centering 
\begin{subtable}{0.8\textwidth} 
\centering 
\label{tab:lndtable} 
\begin{tabular}{|c|c|c|c|} 
\hline Team & Trans Err(mm)$\downarrow$ & Rot Err(deg) $\downarrow$ & ADD Metric(10\%) $\uparrow$\\ 
\hline 
\hline TUDU & 6.38 & 21.33 & 0.13\\ 
IGTUM & \textbf{2.56} & \textbf{5.18} & \bf{0.42}\\ 
EUT & 63.32 & 57.17 & 0.18\\ 
YPL & 44.52 & 51.35 & 0.18\\ 
PVNet & 46.79 & 52.45 & \underline{0.19}\\ 
\hline 
MVL\_3S & \underline{5.91} & 27.21 & 0.12\\ 
\textbf{Ours} & 6.24 & \underline{5.78} & 0.12\\ 
\hline 
\end{tabular} 
\caption{LND} 
\end{subtable} 
\hfill 
\begin{subtable}{0.8\textwidth} 
\centering 
\label{tab:lndocctable} 
\begin{tabular}{|c|c|c|c|} 
\hline Team & Trans Err(mm) $\downarrow$ & Rot Err(deg) $\downarrow$ & ADD Metric(10\%) $\uparrow$\\ 
\hline 
\hline TUDU & \underline{5.83} & 21.48 & 0.11\\ 
IGTUM & \textbf{5.40} & \textbf{10.71} & \textbf{0.37}\\ 
EUT & 66.53 & 28.46 & 0.21\\ 
YPL & 91.91 & 40.82 & 0.26\\ 
PVNet & 28.09 & 17.55 & \underline{0.27} \\ 
\hline 
MVL\_3S & 7.57 & 24.60 & 0.11\\ 
\textbf{Ours} & 8.87 & \underline{13.46} & 0.08\\ 
\hline 
\end{tabular} 
\caption{LND occluded} 
\end{subtable} 

\begin{subtable}{0.8\textwidth} 
\centering 

\label{tab:mbftable} 
\begin{tabular}{|c|c|c|c|} 
\hline Team & Trans Err(mm) $\downarrow$ & Rot Err(deg) $\downarrow$& ADD Metric(10\%) $\uparrow$\\ 
\hline 
\hline TUDU & 5.89 & 17.51 & 0.12\\ 
IGTUM & \textbf{3.00} & \textbf{3.36} & \textbf{0.39}\\ 
EUT & 5.30 & 19.62 & \underline{0.37}\\ 
YPL & 82.87 & 65.60 & 0.15\\ 
PVNet & 3.53 & 25.85 & 0.36\\ 
\hline 
MVL\_3S & \underline{3.45} & 10.39 & 0.35\\ 
\textbf{Ours} & 6.69 & \underline{5.89} & 0.13\\ 
\hline 
\end{tabular} 
\caption{MBF} 
\end{subtable} 
\hfill 
\begin{subtable}{0.8\textwidth} 
\centering 
\label{tab:mbfpcctable} 
\begin{tabular}{|c|c|c|c|} 
\hline Team & Trans Err(mm) $\downarrow$ & Rot Err(deg) $\downarrow$ & ADD Metric(10\%) $\uparrow$\\ 
\hline 
\hline TUDU & 21.50 & 37.93 & 0.04\\ 
IGTUM & \textbf{12.44} & \textbf{18.44} & \textbf{0.27}\\ 
EUT & 80.09 & 39.28 & 0.16 \\ 
YPL & 62.76 & 59.85 & 0.13 \\ 
PVNet & 44.67 & 31.09 & \underline{0.17}\\ 
\hline 
MVL\_3S & 17.55 & 34.13 & 0.11\\ 
\textbf{Ours} & \underline{15.06} & \underline{21.02} & 0.09\\ 
\hline 
\end{tabular} 
\caption{MBF occluded} 
\end{subtable} 
\caption{Table showing comparison of different model performances in EndoVis SurgRIPE challenge across LND, LND occluded, MBF, and MBF occluded datasets. The results of the best-performing model are highlighted in bold, while the second-best model results are underlined, for each type of error. The top five rows show performance for multi-stage methods, while the bottom two rows demonstrate the single-stage methods.}
\label{tab:all_results} 
\end{table}

Table \ref{tab:all_results} summarises the performance of our model on the four test datasets, using translation and rotation errors and the ADD metric. The SurgRIPE dataset was introduced as part of the MICCAI 2022 SurgRIPE challenge, and established a benchmark for 6D surgical instrument pose estimation \cite{surgripechallenge}. The reported results in table \ref{tab:all_results} were obtained from methods evaluated as part of the same challenge, enabling a direct comparison. These methods can broadly be categorised as either multi-stage or single-stage (end-to-end) approaches. Multi-stage methods first predict intermediate representations, such as segmentation masks, keypoints, or 2D-3D correspondences, which are then used in an additional pose estimation or refinement step to obtain the final 6D pose. In contrast, single-stage methods directly predict the 6D pose. The first five methods shown in Table \ref{tab:all_results} use multi-stage approaches, while the final two methods, including our method, use a single-stage, end-to-end approach. \\

The best performing model, by IMF, was an application of SurfEmb, proposed by Haugaard et al. \cite{surfemb}, where dense surface embeddings are learned to establish correspondences between surface points and a canonical model. An iterative render-and-compare strategy was also used to refine an initial pose estimate. Other methods included use of off the shelf networks like SwiftFormer, introduced by Shaker et al. \cite{swiftformer} and EfficientPose, proposed by Bukschat et al. \cite{efficientpose}. Our method displays consistently strong performance, specifically in rotation, while also exhibiting comparable translational performance. Across all four test datasets, our model ranks second in rotational accuracy, with only IMF achieving lower rotation error ($5.78°$ vs. $5.18°$ on LND and $21.02°$ vs. $18.44°$ on MBF occ), demonstrating its robustness even under occlusion. While IMF also achieves the lowest translation error, our model remains competitive, especially in occluded cases where it outperforms most other methods ($8.87$ mm vs. $28.09$ mm for PVNet on LND occ).When compared specifically with the other single-stage method, our model obtains comparable translation errors and significantly lower rotation errors ($5.78°$ vs $27.21°$, $13.46°$ vs $24.60°$, $5.89°$ vs $10.39°$ and $21.02°$ vs $34.13°$ for LND, LND occluded, MBF and MBF occluded datasets, respectively). While our method achieves good rotational and translational accuracy, it attains an ADD metric of $0.12$, $0.08$, $0.13$, and $0.09$ for the  LND, LND occluded, MBF and MBF occluded datasets, respectively. These were the lowest of all methods, including those with highest rotation and translation errors, such as YPL ($0.18$, $0.26$, $0.15$, $0.13$). We examine the source of this discrepancy in Section \ref{sec:error-analysis}
\subsection{Runtime Analysis}
We evaluate the computational efficiency of the model on a consumer-level NVIDIA Tesla T4 GPU. The end-to-end inference time of our proposed pipeline was 33.3 frames per second corresponding to approximately 30ms per image. In comparison, PVNet \cite{pvnet}, a two-stage pose estimation framework, reports an inference time of approximately 25 FPS on a GTX 1080ti GPU. As real-time is generally considered to be over 30 FPS, this demonstrates that the PICO achieves superior real-time performance, alongside lower translation and rotational errors in most test cases.

\subsection{Ablation Studies}
Table \ref{tab:ablation} shows ablation studies that evaluate the inclusion of terms in the multi-task loss function across all 4 test datasets (LND, LND Occluded, MBF and MBF Occluded). A comparison of model performances when trained with only translation and rotation terms (top row), then translation and rotation with segmentation and depth terms added separately and together (middle rows) then finally all loss terms (bottom row) is shown in Tables 2a to 2d, with the lowest errors shown in bold. The auxiliary tasks constrain different pose components. Depth supervision yields the lowest translation error on LND, LND occluded, and MBF (6.22, 8.36, and 5.27\,mm), primarily constraining the $z$ component. Segmentation improves rotation on LND (18.89 to 11.11) and MBF (12.95 to 9.67) by encoding the tool's projected shape. However, their combination does not accumulate these gains: on MBF, it performs worse than either task alone on both metrics, and on MBF occluded, worse than the baseline in translation. As projection and point-to-point losses couple both signals through a single predicted transformation, the full loss resolves this trade-off, improving both metrics over the segmentation--depth configuration on LND, MBF, and MBF occluded, and translation on LND occluded.

\begin{table}[!h] 
\centering 
\begin{subtable}{0.8\textwidth} 
\centering 
\begin{tabular}{|l|c|c|} 
\hline Loss & Trans Err(mm) $\downarrow$& Rot Err(deg) $
\downarrow$ \\ 
\hline 
Translation + Rotation & 7.49 & 18.89 \\ 
+ Seg & 9.98 & 11.11 \\
+ Depth & \textbf{6.22} & 11.06 \\
+ Seg + Depth & 7.31 & 10.24 \\
+ Points + Projection & 6.24 & \textbf{5.78} \\ 
\hline 
\end{tabular} 
\caption{LND} 
\label{tab:lndab} 
\end{subtable} 
\hfill 
\begin{subtable}{0.8\textwidth} 
\centering 
\begin{tabular}{|l|c|c|} 
\hline Loss & Trans Err(mm) $\downarrow$ & Rot Err(deg) $\downarrow$  \\ 
\hline 
Translation + Rotation & 13.28 & 11.63 \\ 
+ Seg & 10.05 & 17.61 \\
+ Depth & \textbf{8.36} & 14.31 \\
+ Seg + Depth & 11.93 & \textbf{10.74} \\
+ Points + Projection & 8.87 & 13.46 \\ 
\hline 
\end{tabular} 
\caption{LND occluded} 
\label{tab:lndoccab} 
\end{subtable} 

\begin{subtable}{0.8\textwidth} 
\centering 
\begin{tabular}{|l|c|c|} 
\hline Loss & Trans Err(mm) $\downarrow$ & Rot Err(deg) $\downarrow$ \\ 
\hline 
Translation + Rotation & 9.01 & 12.95 \\ 
+ Seg & 7.41 & 9.67 \\
+ Depth & \textbf{5.27} & 10.54 \\
+ Seg + Depth & 8.19 & 12.32 \\
+ Points + Projection & 6.69 & \textbf{5.89} \\ 
\hline 
\end{tabular} 
\caption{MBF} 
\label{tab:mbfab} 
\end{subtable} 
\hfill 
\begin{subtable}{0.8\textwidth} 
\centering 
\begin{tabular}{|l|c|c|} 
\hline Loss  & Trans Err(mm) $\downarrow$ & Rot Err(deg) $\downarrow$\\ 
\hline 
Translation + Rotation & 16.31 & 28.92 \\ 
+ Seg & \textbf{13.69} & 26.71 \\
+ Depth & 15.84 &  23.54 \\
+ Seg + Depth & 17.01 & 24.98 \\
+ Points + Projection & 15.06 & \textbf{21.02} \\ 
\hline 
\end{tabular} 
\caption{MBF occluded} 
\label{tab:mbfoccab} 
\end{subtable} 
\caption{Table showing ablation studies for multi-task loss function and impact of of auxiliary tasks on model performance, evaluated across EndoVis SurgRIPE challenge LND, LND occluded, MBF, and MBF occluded datasets.} 
\label{tab:ablation} 
\end{table}

\subsection{Error Decomposition and the ADD Metric}\label{sec:error-analysis}

\begin{figure}[ht]
\centering

\begin{subfigure}[t]{0.24\textwidth}
    \centering
    \includegraphics[width=\textwidth]{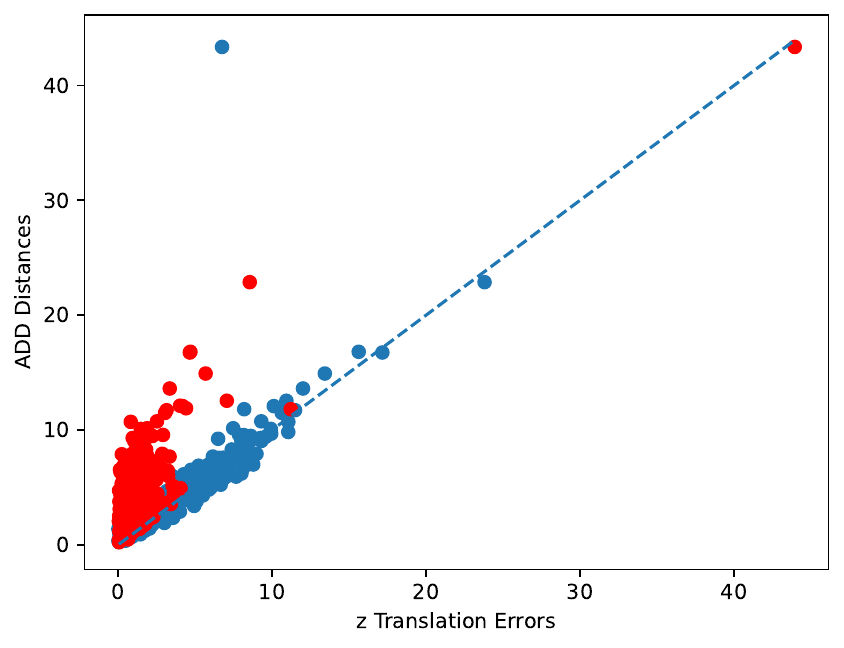}
    \caption{LND}
    \label{fig:lnd}
\end{subfigure}
\hfill
\begin{subfigure}[t]{0.24\textwidth}
    \centering
    \includegraphics[width=\textwidth]{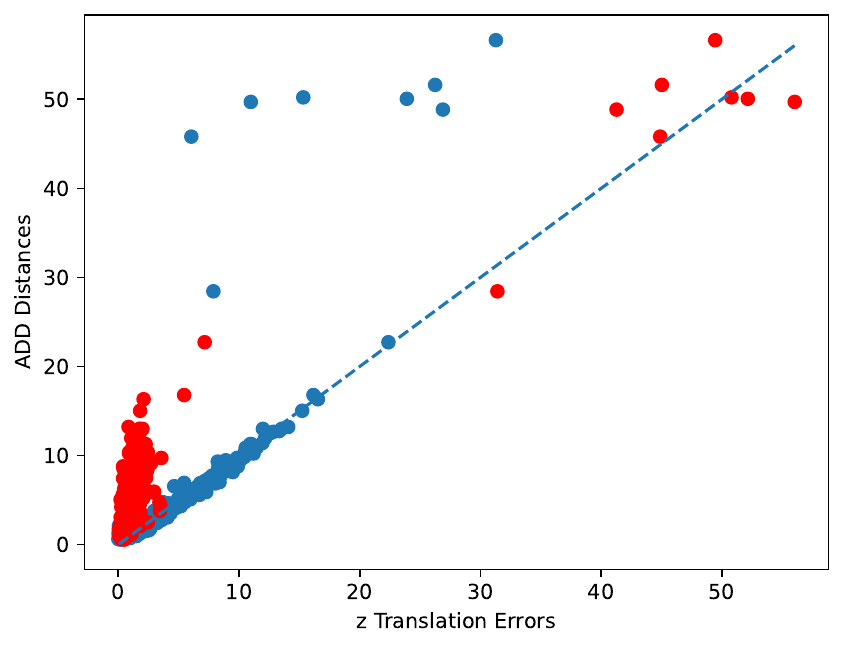}
    \caption{LND Occluded}
    \label{fig:lnd_occ}
\end{subfigure}
\hfill
\begin{subfigure}[t]{0.24\textwidth}
    \centering
    \includegraphics[width=\textwidth]{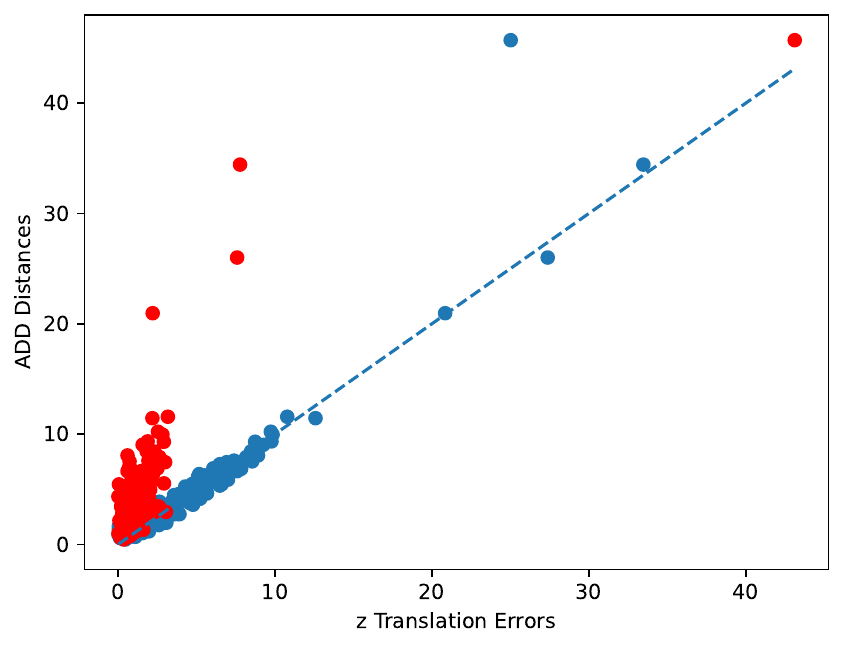}
    \caption{MBF}
    \label{fig:mbf}
\end{subfigure}
\hfill
\begin{subfigure}[t]{0.24\textwidth}
    \centering
    \includegraphics[width=\textwidth]{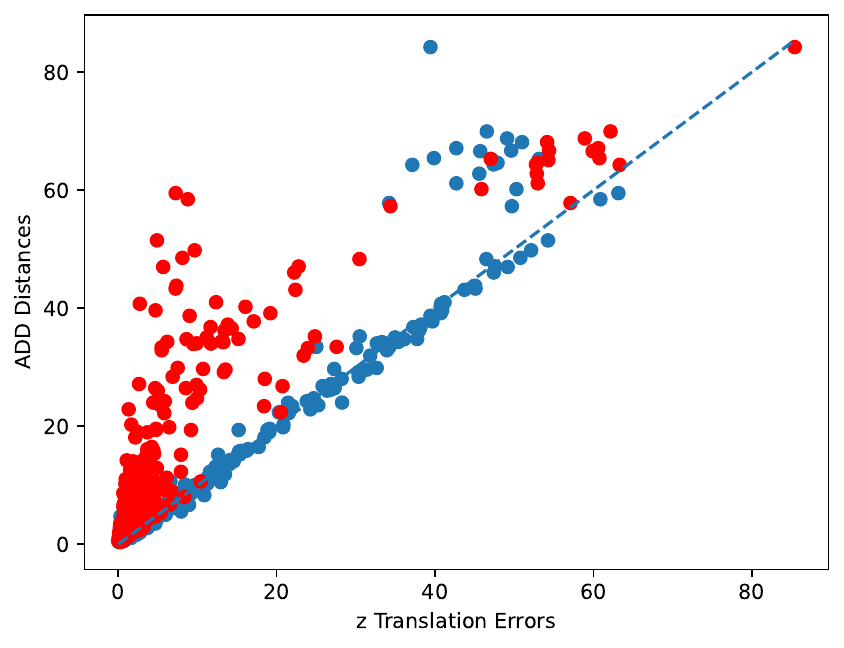}
    \caption{MBF Occluded}
    \label{fig:mbf_occ}
\end{subfigure}

\caption{Relationship between $xy$ and $z$ translation errors and ADD distances for the four datasets. Each plot contains both $xy$ (red) and $z$ (blue) translation error measurements plotted against ADD distance. The dashed line represents $x=y$, indicating where the translation error and ADD distance are equal.}
\label{fig:error-adds-comparison}

\end{figure}

Our model achieves lower ADD scores than methods with larger translation and rotation errors, suggesting that ADD captures a failure mode not reflected by either metric alone. We therefore decompose translation error into image-plane $(x,y)$ and depth $(z)$ components and compare them with per-sample ADD distances (Fig \ref{fig:error-adds-comparison}).  Across all test sets, $z$ error, shown in Figure \ref{fig:error-adds-comparison} by the blue points, is larger than $xy$ error, shown in red, with mean values of 3.88, 6.09, 3.90 and 11.82mm compared with 2.68, 1.17, 1.34 and 5.82mm, for LND, LND Occluded, MBF and MBF Occluded datasets, respectively. More importantly, $z$ error closely tracks ADD distance, exhibiting a strong monotonic relationship, with Spearman correlations of 0.955, 0.978, 0.963 and 0.982, compared with 0.449, 0.650, 0.602 and 0.847 for $xy$ error. These consistently higher Spearman correlations for $z$ indicate that ADD distance is sensitive to the magnitude of depth errors. Therefore depth error, which displaces the entire point cloud along the camera axis, is the primary limiting factor for ADD.

\section{Discussion}
Our results show that the PICO framework provides robust and accurate 6DoF surgical tool pose estimation, even under occlusion. Across all test sets, it achieves competitive translation and rotation accuracy, highlighting the value of combining multi-task learning with spatially aware proxy tasks. Our ablations show that the auxiliary tasks act on different components of the pose, depth on translation and segmentation on orientation, but that they interfere when optimised jointly. The projection and point-to-point losses, computed from the predicted pose, couple both signals to a single transformation and recover this lost performance, accounting for the improvement of the full model. This decomposition identifies depth estimation as the principal limitation on our ADD scores, and the ablation confirms that the auxiliary depth task attacks it directly, reducing translation error from 13.28 to 8.36 mm on LND occluded. Closing the remaining gap requires depth supervision more accurate than our current pseudo-depth maps provide. Compared to the other single-stage method, PICO substantially improves rotational accuracy without compromising translation. It remains competitive with multi-stage frameworks, suggesting that end-to-end single-stage networks can match iterative refinement approaches when guided by geometric consistency losses. Furthermore, PICO operates with a runtime of $\approx$33 FPS on a consumer-level NVIDIA Tesla T4 GPU, demonstrating suitability for real-time surgical applications.

\noindent One limitation of the proposed approach is that resizing non-square image crops to a fixed square resolution may introduce minor aspect ratio distortions. This effect is mitigated by the tight centring of the crop around the detected tool, which preserves relevant geometric structure. Another limitation is the pseudo-depth generation method. Randomly sampling points from the mesh and assigning them to the nearest pixel only provides an approximation of hte true visible surface, while filling holes using neighbouring depth values could smooth object boundaries and potentially introduce geometrically unrealistic depth estimates. In addition, due to the limited availability of datasets with ground-truth pose annotations in the surgical domain, SurgRIPE remains the only one publicly available. Consequentially, comparison with competing methods are restricted to results reported in the corresponding benchmark challenge paper \cite{surgripechallenge}, rather than fully reproducible implementations.\\
Challenges remain in generalising to unseen instruments or surgical environments, indicating the need for data augmentation and domain adaptation.

\section{Conclusion}\label{}
In this work, we address the problem of accurate and robust 6DoF surgical tool pose estimation from monocular RGB images, where occlusion and limited geometric information make spatial reasoning challenging. We proposed the PICO model, a novel real-time end-to-end trainable approach to 6DoF surgical tool pose estimation. By leveraging multi-task learning, our model jointly predicts segmentation and depth while regressing rotation and translation parameters. We also introduced two proxy tasks - projection and point-to-point tasks - that enforce geometric consistency in 2D and 3D spaces, improving the robustness and accuracy of pose estimation. 

\noindent Experimental results on the SurgRIPE dataset demonstrate that PICO achieves competitive translational and rotational accuracy to existing methods, especially in occluded settings, which underlines its robustness and highlights the importance of incorporating geometric consistency in the learning process. Notably, the proposed single-stage framework significantly improves rotational accuracy in comparison to the only other single-stage method (MVL\_3S), while also remaining competitive to multi-stage methods and operating with a runtime of $\approx$33 FPS, demonstrating suitability for real-time surgical applications.

\noindent Despite these encouraging results, challenges remain. In particular, poor depth estimation performance limits our performance on the ADD metric. Additionally, preprocessing-related distortions and few publicly released surgical pose datasets limits evaluation and reproducibility.
Further improvements will focus on depth modelling and enhancing generalisation to unseen tools, surgical environments and scenarios through domain adaptation and data augmentation. Overall, PICO demonstrates a promising path towards real-time, markerless 6DoF pose estimation for surgical robotics.

\backmatter

\section*{Statements and Declarations}

\textbf{Acknowledgements} This paper was completed as part of the UKRI Centre for Doctoral Training in Artificial Intelligence for Medical Diagnosis and Care at University of Leeds, funded by the EPSRC grant number [EP/Y009819/1]

\noindent \textbf{Competing Interests} The authors have no relevant conflicts of interest to declare.

\noindent

\bibliography{sn-bibliography}%

\end{document}